\documentclass{article}
\usepackage{spconf,amsmath,graphicx,hyperref}
\usepackage{etoolbox}
\apptocmd{\thebibliography}
  {%
    \fontsize{9pt}{11pt}\selectfont
    \setlength{\itemsep}{0pt}
    \setlength{\parsep}{0pt}
    \setlength{\parskip}{0pt}
  }
  {}
  {}
\usepackage{booktabs}
\usepackage{amsmath,amssymb,amsfonts}
\usepackage{indentfirst}
\usepackage{enumitem}

\newcommand{\method}{MDN-Control}
\newcommand{\clipt}{CLIP-T}
\newcommand{\clipf}{CLIP-F}
\newcommand{\cmerr}{CM-Err}
\newcommand{\warperr}{Warp-Err}

\title{MDN-Control: Mask-Depth-Noise Guided Region Control for Multi-Subject Video Editing}

\name{Jiayi Yu \qquad Xi Ye \qquad Lina Wang$^{\dagger}$\thanks{%
\fontsize{9pt}{11pt}\selectfont
$\dagger$ Corresponding author}
\qquad Yunkun Xia}
\address{School of Cyber Science and Engineering, Wuhan University}

\usepackage{eso-pic}

\AddToShipoutPictureFG*{%
  \AtPageLowerLeft{%
    \put(0,\LenToUnit{10mm}){%
      \makebox[\paperwidth][c]{%
        \parbox[b]{0.9\paperwidth}{%
          \centering
          \normalfont
          \fontsize{8pt}{9pt}\selectfont
          This work has been submitted to the IEEE for possible publication.\\
          Copyright may be transferred without notice, after which this version
          may no longer be accessible.\par
        }%
      }%
    }%
  }%
}

\begin{document}
%
\maketitle
\setlength{\parskip}{0pt}
\setlength{\abovedisplayskip}{5pt plus 1pt minus 1pt}
\setlength{\belowdisplayskip}{5pt plus 1pt minus 1pt}
\setlength{\abovedisplayshortskip}{3pt plus 1pt minus 1pt}
\setlength{\belowdisplayshortskip}{3pt plus 1pt minus 1pt}

\setlength{\floatsep}{1pt}

\setlength{\abovecaptionskip}{0pt}
\setlength{\belowcaptionskip}{0pt}

\setlength{\textfloatsep}{6pt}

\setlength{\dbltextfloatsep}{6pt}

\setlength{\intextsep}{8pt plus 2pt minus 2pt}

\begin{abstract}
Multi subject video editing modifies designated subjects while preserving non target content, but faces cross subject attribute leakage, and occlusion ambiguity. Existing approaches rely on masks and struggle to distinguish overlapping subjects or ensure consistent generation. To address these limitations, we propose \method{}, a training free framework jointly controlling target localization, occlusion geometry, and appearance initialization. Specifically, mask-guided localization provides consistent target localization, while depth-aware occlusion control resolves ambiguous boundaries between overlapping subjects. We further introduce noise latent prompting, which retrieves Gaussian initializations from a noise library for prompt relevant priors. Experiments on MSVBench show that \method{} achieves the lowest CM-Err and the highest Q-Edit, while maintaining competitive text alignment and temporal consistency, demonstrating the effectiveness of combining spatial, geometric, and latent priors for multi subject video editing.  
\end{abstract}

\section{Introduction}
\label{sec:intro}
Recent diffusion models have advanced visual generation and text guided video editing~\cite{ho2020ddpm,ho2022video,blattmann2023align}. Video editing methods improve content preservation and temporal consistency through model adaptation, attention control, and feature propagation~\cite{wu2023tuneavideo,qi2023fatezero,geyer2023tokenflow}. Other studies introduce depth, motion, and compositional conditions to preserve source structure~\cite{esser2023gen1,wang2023videocomposer}, while recent approaches support localized editing with spatial conditions and pretrained video priors~\cite{jiang2025vace,shen2025astra}. Conditional generation studies show that explicit spatial and subject guidance helps preserve geometric structure and subject attributes~\cite{shen2024imagpose}. These advances make it practical to modify designated subjects while preserving unrelated content.

\begin{figure}[!t]
    \centering
    \includegraphics[width=0.82\columnwidth]{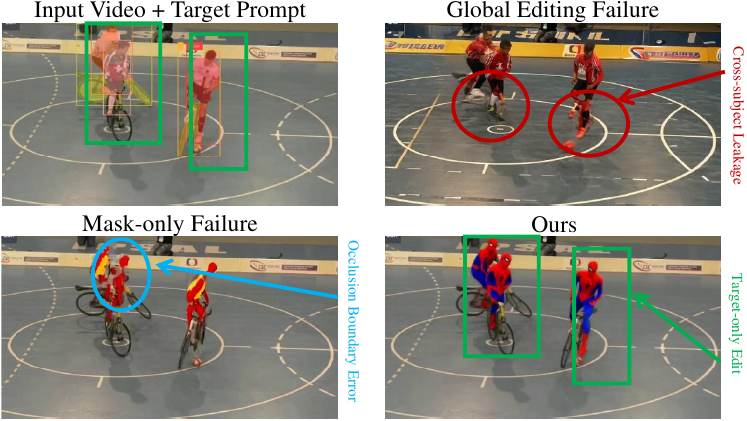}
    \vspace{-0.3cm}
    \caption{Motivation of multi-subject target-region video editing. The designated subject should be edited despite occlusion and neighboring-subject interference, while non-target subjects and the background remain unchanged.}
    \label{fig:motivation}
\end{figure}

However, precise control remains difficult in multi subject videos, especially when subjects are adjacent or occluded. As illustrated in Fig.~\ref{fig:motivation}, an edit intended for one subject can unintentionally affect neighboring subjects, resulting in target drift and cross subject attribute leakage. Masks provide spatial constraints, but tracking errors and uncertain boundaries can degrade localization~\cite{ravi2024sam2}. More importantly, binary masks do not encode the front and back ordering of overlapping subjects. Even accurate localization does not guarantee that the generated appearance is well aligned with the editing prompt, since different initial noise latents can lead to different visual outcomes. Therefore, robust multi subject editing requires coordinated control over target localization, occlusion geometry, and appearance initialization.

To address these challenges, we propose Mask-Depth-Noise Guided Region Control (\method{}), a training free framework for controllable multi subject video editing. First, text guided detection and video mask tracking localize the designated subject throughout the video. Second, depth cues are introduced within the target neighborhood to resolve ambiguous occlusion boundaries and distinguish overlapping subjects. Third, we introduce noise latent prompting, which retrieves Gaussian initializations from an offline noise appearance library according to the similarity between their recorded generation outcomes and the requested appearance. These three components provide complementary spatial, geometric, and latent guidance without updating the generation model.
Our contributions are summarized as follows:
\begin{itemize}[nosep, leftmargin=1.51em, labelsep=0.6em]
    \item We formulate multi subject video editing as a control problem involving target localization, occlusion geometry, and appearance initialization, and propose a training free framework that addresses these factors jointly.

    \item We combine mask based localization with depth guided occlusion reasoning to reduce target drift and cross subject interference, especially around ambiguous boundaries between overlapping subjects.

    \item We introduce noise latent prompting, which retrieves prompt relevant Gaussian initializations from an offline noise appearance library to provide controllable appearance priors without parameter updates.

    \item Experiments on MSVBench demonstrate that \method{} achieves the lowest \cmerr{} and the highest Q Edit among the evaluated methods, while maintaining competitive text alignment and temporal consistency.
\end{itemize}

\begin{figure*}[!t]
    \centering
    \includegraphics[width=0.98\linewidth]{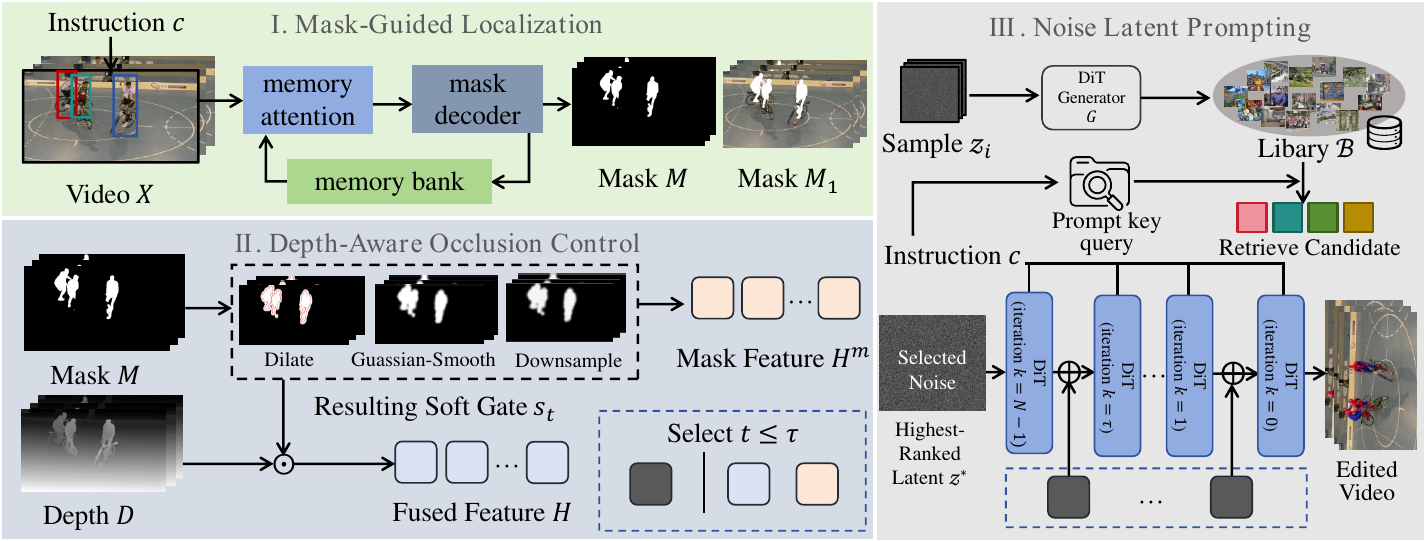}
    \vspace{-0.3cm}
    \caption{Overview of \method{}. Mask-guided localization localizes the target, depth-aware occlusion control guides editing near occlusions, and noise latent prompting initializes appearance generation. Denoising uses mask-depth guidance at early steps and mask-only guidance thereafter.}
      \vspace{-0.3cm}
    \label{fig:framework}
\end{figure*}

\section{Methodology}
\label{sec:format}

As illustrated in Fig.~\ref{fig:framework}, \method{} combines mask-guided localization, depth-aware occlusion control, and noise latent prompting. These components specify where to edit, provide local visibility cues, and initialize prompt-relevant appearance generation, respectively. All models remain frozen during editing.

\noindent\textbf{Overview.}
Given a video $X=\{x_t\}_{t=1}^{T}$ and an instruction $c$ specifying the target subject and desired appearance, we seek an edited video $\hat{X}=\{\hat{x}_t\}_{t=1}^{T}$ satisfying
\begin{equation}
\begin{aligned}
m_t\odot\hat{x}_t &\models c,\\
(1-m_t)\odot\hat{x}_t &\approx (1-m_t)\odot x_t,
\end{aligned}
\label{eq:task_definition}
\end{equation}
where $m_t$ is the target mask and $\hat{x}_t$ is the edited frame. The first condition expresses target-text alignment, while the second requires preservation of non-target subjects and the background. Mask-guided localization restricts depth guidance to the target neighborhood, and retrieved noise initializes appearance generation under these spatial constraints.

\noindent\textbf{Mask-Guided Localization.}
We first apply a text-guided detector~\cite{liu2023groundingdino} to the initial frame $x_1$ and instruction $c$ to obtain the target box $b_1$. The detector selects visual queries through their similarity to text features and decodes the queries into a bounding box. The box prompts a video segmentation model~\cite{ravi2024sam2}, which initializes the target mask and propagates it across frames through memory attention. The resulting sequence $M=\{m_t\}_{t=1}^{T}$ provides instance-specific spatial support without repeating text-guided detection in every frame. This support also restricts subsequent depth guidance to the target neighborhood, reducing interference from unrelated subjects. However, masks alone do not explicitly represent the visibility ordering of overlapping instances, motivating the following depth-aware occlusion control.

\noindent\textbf{Depth-Aware Occlusion Control.}
We estimate depth maps $D=\{d_t\}_{t=1}^{T}$ using a pretrained depth estimator~\cite{yang2024depthanythingv2}. To accommodate uncertain mask boundaries, each binary mask is dilated, Gaussian-smoothed, and downsampled to the conditioning feature resolution:
\begin{equation}
s_t=\operatorname{Down}\!\left(
G_{\sigma}*\operatorname{Dilate}_{r}(m_t)
\right),
\end{equation}
where $r$ is the dilation radius and $\sigma$ controls smoothing. The resulting soft gate $s_t\in[0,1]$ combines aligned mask features $h_t^m$ and depth features $h_t^d$:
\begin{equation}
h_t=s_t\odot h_t^d+(1-s_t)\odot h_t^m.
\end{equation}
Depth features provide front--back cues within the softened target neighborhood, while mask features dominate elsewhere. Local gating associates geometric information with the selected instance, since depth alone does not identify which subject should be edited.

We use fused guidance during early denoising and mask-only guidance during later refinement, following the coarse-to-fine behavior of diffusion generation~\cite{wu2024freediff}:
\begin{equation}
C_k=
\begin{cases}
H, & 0\leq k\leq\tau,\\
H^m, & \tau<k< N,
\end{cases}
\end{equation}
where $k$ counts denoising iterations sequentially, $N$ is the step count, $H=\{h_t\}_{t=1}^{T}$, $H^m=\{h_t^m\}_{t=1}^{T}$, and $\tau$ is the switching step. We apply geometric cues during structural formation while retaining spatial constraints for detail refinement.

\noindent\textbf{Noise Latent Prompting.}
Spatial guidance determines where to edit but does not uniquely determine the appearance. Different initial noise latents produce different shapes, colors, and textures under the same conditions~\cite{ge2023correlation,wang2025noisequery}. Independent Gaussian sampling is prompt-agnostic, and finding a suitable appearance through repeated generation incurs additional inference cost. Therefore, we evaluate candidate noises offline and construct a library that associates each initialization with its observed generation outcome.
We sample $z_i\sim\mathcal{N}(0,I)$ and generate $y_i=G(z_i,c_0)$ using the frozen DiT~\cite{jiang2025vace} generator $G$ under an empty condition $c_0$. The library stores:
\begin{equation}
\mathcal{B}=\{(z_i,\phi_i)\}_{i=1}^{L},
\qquad \phi_i=\Phi(y_i),
\end{equation}
where $L$ is the library size and $\Phi$ extracts CLIP image embeddings~\cite{radford2021clip} and low-level color and texture descriptors. For an editing instruction $c$, the query $\psi(c)$ combines its CLIP text embedding with the requested appearance attributes. We retrieve candidate indices by
\begin{equation}
\mathcal{R}_K(c)=
\operatorname{TopK}_{i}\,S\!\left(\psi(c),\phi_i\right),
\label{eq:noise_retrieval}
\end{equation}
where $S$ measures CLIP cosine similarity, low-level attribute similarity and their weighted combination. We further select the highest-ranked latent $z^*$ for an edit or candidates from $\mathcal{R}_K(c)$ for appearance variation. Retrieval uses recorded outputs and does not guarantee superiority over every random seed. The latent initializes generation, while the text instruction and $C_k$ guide denoising, complementing mask and depth control without parameter updates.

\section{Experiments and Analysis}\label{sec:exp}
\label{sec:pagestyle}

\noindent\textbf{Implementation Details.}
We evaluate \method{} on MSVBench~\cite{shen2025astra} against FateZero~\cite{qi2023fatezero}, TokenFlow~\cite{geyer2023tokenflow}, Video\-Painter~\cite{bian2025videopainter}, VideoGrain~\cite{yang2025videograin}, DMT~\cite{yatim2024spacetime}, and ASTRA~\cite{shen2025astra}. Experiments use one NVIDIA L40 GPU, Grounded SAM 2 masks~\cite{ren2024groundedsam,ravi2024sam2}, and Depth Anything V2 maps~\cite{yang2024depthanythingv2}. Metrics include \warperr{} ~\cite{teed2020raft} for background temporal consistency, \clipt{} ~\cite{radford2021clip} for text alignment with non-target pixels blacked out, \clipf{}~\cite{radford2021clip} for adjacent-frame CLIP similarity, \cmerr{}~\cite{shen2025astra} for multi-subject layout preservation, and Q-Edit~\cite{cong2024flatten} for mean \clipt{} divided by mean \warperr{}.

\noindent\textbf{Quantitative Comparison.}
As illustrated in table~\ref{tab:main_results}, \method{} obtains the lowest mean \warperr{} (2.87), highest mean Q-Edit (9.43), and lowest mean \cmerr{} (2.75), ranking second in \clipt{} and \clipf{}, which demonstrates a favorable balance between temporal consistency, text alignment, frame consistency and multi-subject layout preservation.
\begin{table}[t]
\centering
\caption{Quantitative comparison on MSVBench.}
\label{tab:main_results}
\fontsize{9pt}{11pt}\selectfont
\setlength{\tabcolsep}{1pt}
\begin{tabular}{@{}lccccc@{}}
\toprule
Method & \warperr{}$\downarrow$ & \clipt{}$\uparrow$
& \clipf{}$\uparrow$ & Q-Edit$\uparrow$ & \cmerr{}$\downarrow$ \\
\midrule
FateZero~\cite{qi2023fatezero}
& 3.53 & 25.28 & \textbf{95.86} & 7.17 & 2.95 \\
TokenFlow~\cite{geyer2023tokenflow}
& 7.79 & 25.58 & 93.69 & 3.28 & 4.26 \\
VideoPainter~\cite{bian2025videopainter}
& 4.63 & 25.48 & 93.90 & 5.51 & 3.95 \\
VideoGrain~\cite{yang2025videograin}
& 7.87 & 26.98 & 95.19 & 3.43 & 3.19 \\
DMT~\cite{yatim2024spacetime}
& 5.77 & 26.35 & 95.34 & 4.56 & 4.30 \\
ASTRA~\cite{shen2025astra}
& \underline{2.94} & \textbf{27.49} & 95.60
& \underline{9.34} & \underline{2.78} \\
\method{}
& \textbf{2.87} & \underline{27.09} & \underline{95.72}
& \textbf{9.43} & \textbf{2.75} \\
\bottomrule
\end{tabular}
\end{table}

\begin{figure*}[!t]
    \centering
    \includegraphics[width=0.95\linewidth]{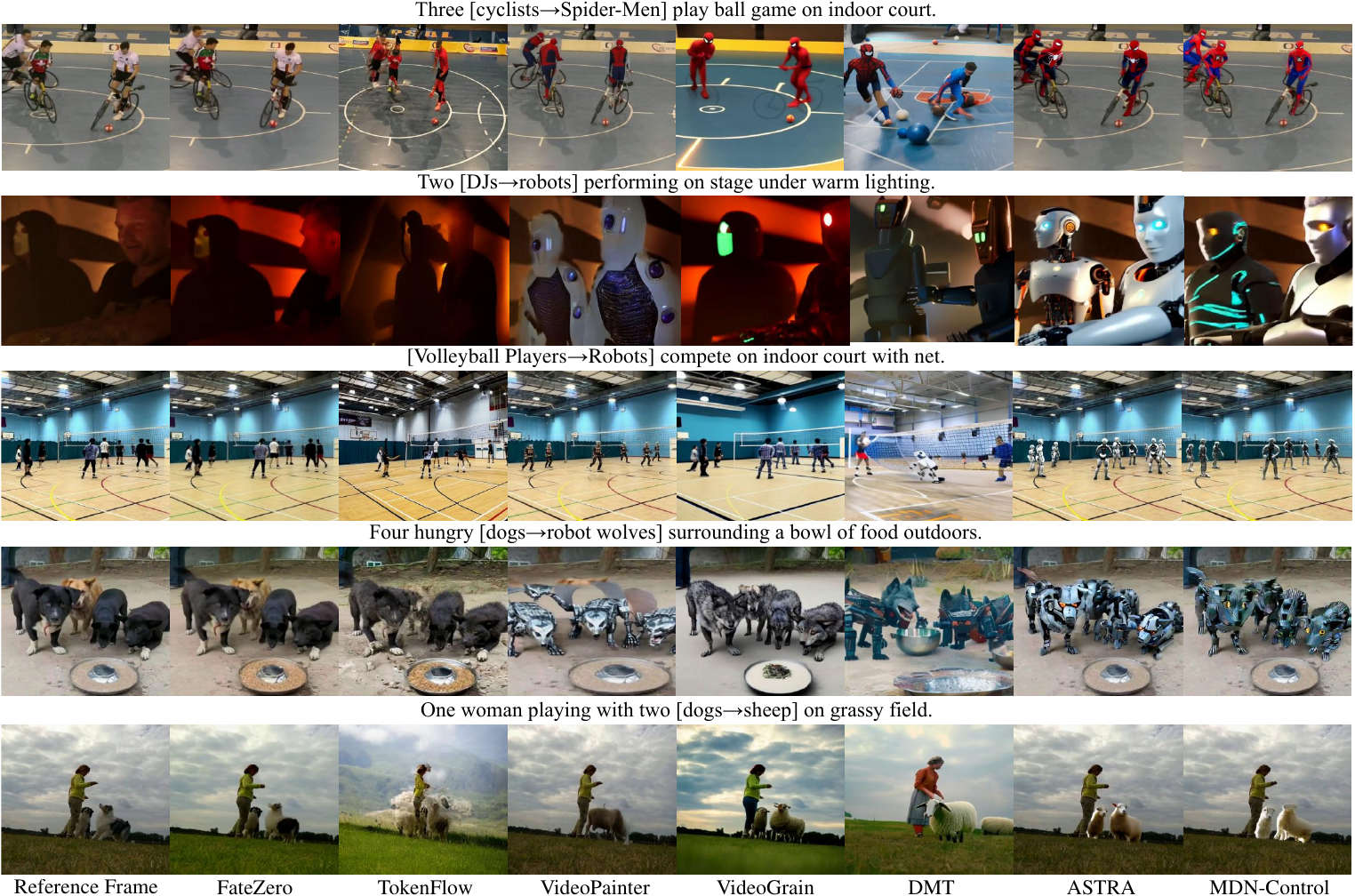}
       \vspace{-0.3cm}
    \caption{Qualitative comparison on MSVBench. The proposed method better edits the target subject while preserving neighboring subjects and background regions in crowded and occluded scenes.}
    \vspace{-0.3cm}
    \label{fig:qualitative}
\end{figure*}

\noindent\textbf{Qualitative Results.}
As illustrated in Fig.~\ref{fig:qualitative}, most baselines suffer from incomplete editing, background damage, or cross-subject attribute leakage. \method{} edits the designated subject while keeping non-target regions nearly unchanged, illustrating the effectiveness of jointly combining mask, depth, and noise guidance for target-specific editing with plausible boundaries and prompt-aligned appearances.

\noindent\textbf{User Study.}
We conduct a user study to evaluate subjective generation quality. For each method, we randomly select 20 videos and organize them into 20 comparison groups. Each group contains the results from \method{} and the six baseline methods for the same editing case. As shown in Table~\ref{tab:user_study}, \method{} obtains the highest scores in text alignment and subject diversity, and achieves competitive performance in background preservation and video quality. 

\begin{table}[t]
\centering
\caption{User study results. BG: background preservation;
TA: text alignment; SD: subject diversity; VQ: video quality.}
\label{tab:user_study}
\fontsize{9pt}{11pt}\selectfont
\setlength{\tabcolsep}{2pt}
\begin{tabular*}{\columnwidth}{@{\extracolsep{\fill}}lcccc@{}}
\toprule
Method & BG & TA & SD & VQ \\
\midrule
FateZero~\cite{qi2023fatezero}
& 1.20 & 2.00 & 2.00 & 1.70 \\
TokenFlow~\cite{geyer2023tokenflow}
& 3.00 & 2.30 & 4.00 & 3.10 \\
VideoPainter~\cite{bian2025videopainter}
& 3.20 & 2.70 & 3.50 & 3.30 \\
VideoGrain~\cite{yang2025videograin}
& 3.80 & 3.30 & 3.00 & 3.60 \\
DMT~\cite{yatim2024spacetime}
& 4.20 & 4.10 & \underline{4.10} & 4.40 \\
ASTRA~\cite{shen2025astra}
& \textbf{5.00} & \underline{4.80} & 3.60 & \textbf{5.20} \\
\method{}
& \underline{4.90} & \textbf{4.90} & \textbf{5.20}
& \underline{5.10} \\
\bottomrule
\end{tabular*}
\end{table}

\noindent\textbf{Ablation Study.}
We evaluate seven module combinations on MSVBench.
Base denotes the base DiT generator; M and D denote mask-guided
localization and depth guidance, respectively. Rnd. denotes fixed
Gaussian initialization, and Ret. denotes retrieved noise.
M+D+Rnd. combines both spatial controls, while M+D+Ret. is the
full model, \method{}.
\begin{table}[t]
\centering
\caption{Module ablation on MSVBench.}
\label{tab:ablation}
\label{tab:extended_modules}
\fontsize{9pt}{11pt}\selectfont
\setlength{\tabcolsep}{1pt}
\begin{tabular*}{\columnwidth}{@{\extracolsep{\fill}}lccccc@{}}
\toprule
Config.
& \warperr{}$\downarrow$
& \clipt{}$\uparrow$
& \clipf{}$\uparrow$
& Q-Edit$\uparrow$
& \cmerr{}$\downarrow$ \\
\midrule
Base
& 2.71 & 24.33 & \underline{96.13} & 8.98 & 3.21 \\
M
& 2.91 & \textbf{27.16} & 95.79 & 9.33 & 3.17 \\
D
& \textbf{2.71} & 24.33 & 96.12 & 8.99 & 3.28 \\
M+Ret.
& 2.91 & \underline{27.11} & 95.63 & 9.31 & 3.23 \\
D+Ret.
& \underline{2.71} & 24.33 & \textbf{96.13} & 8.99 & 3.22 \\
M+D+Rnd.
& 2.87 & 27.00 & 95.60 & \underline{9.41} & \textbf{2.69} \\
M+D+Ret.
& 2.87 & 27.09 & 95.72 & \textbf{9.43} & \underline{2.75} \\
\bottomrule
\end{tabular*}
\end{table}
Table~\ref{tab:ablation} shows complementary effects: mask guidance improves text alignment, while depth improves layout preservation. Retrieval further raises Q-Edit but slightly increases \cmerr{}, possibly reflecting a trade-off between appearance matching and layout preservation. The full model achieves the highest Q-Edit and lower \cmerr{} than either spatial component alone, supporting the benefit of joint control.

\noindent\textbf{Depth Contribution.}
We compare outputs generated with mask and retrieved noise, with and without depth guidance, to examine the effect of geometric conditioning.
\begin{table}[!t]
\centering
\caption{Comparison of depth guidance. Both variants use mask guidance and retrieved noise.}
\label{tab:depth_ablation}
\fontsize{9pt}{11pt}\selectfont
\setlength{\tabcolsep}{1pt}
\begin{tabular*}{\columnwidth}{@{\extracolsep{\fill}}lccccc@{}}
\toprule
Config.
& \warperr{}$\downarrow$
& \clipt{}$\uparrow$
& \clipf{}$\uparrow$
& Q-Edit$\uparrow$
& \cmerr{}$\downarrow$ \\
\midrule
w/o Depth
& 2.91 & \textbf{27.11} & 95.63 & 9.31 & 3.23 \\
w/ Depth
& \textbf{2.87} & 27.09 & \textbf{95.72}
& \textbf{9.43} & \textbf{2.75} \\
\bottomrule
\end{tabular*}
\par
\normalsize
\caption{Comparison of fixed Gaussian seeds and retrieved initialization on MSVBench.}
\label{tab:fixed_noise}
\fontsize{9pt}{11pt}\selectfont
\setlength{\tabcolsep}{1pt}
\begin{tabular*}{\columnwidth}{@{\extracolsep{\fill}}lccccc@{}}
\toprule
Noise
& \warperr{}$\downarrow$
& \clipt{}$\uparrow$
& \clipf{}$\uparrow$
& Q-Edit$\uparrow$
& \cmerr{}$\downarrow$ \\
\midrule
Seed 0
& \underline{2.87} & \textbf{27.12} & \textbf{95.79}
& \textbf{9.46} & 2.74 \\
Seed 1
& 2.89 & 27.00 & 95.61 & 9.34 & 2.75 \\
Seed 42
& \underline{2.87} & 26.97 & 95.68
& 9.40 & \underline{2.67} \\
Seed 1234
& \textbf{2.86} & 27.03 & \underline{95.72}
& \underline{9.45} & \textbf{2.65} \\
Seed 2025
& 2.89 & \underline{27.10} & 95.60
& 9.39 & \underline{2.67} \\
\midrule
Fixed avg.
& 2.88 & 27.05 & 95.68 & 9.41 & 2.70 \\
\method{}
& \underline{2.87} & 27.09 & \underline{95.72}
& 9.43 & 2.75 \\
\bottomrule
\end{tabular*}
\end{table}
As shown in Table~\ref{tab:depth_ablation}, the depth-enabled outputs have lower \warperr{} and \cmerr{} and higher \clipf{} and Q-Edit, indicating that depth guidance improves temporal coherence and multi-subject spatial preservation, with a minor trade-off in text alignment.

\noindent\textbf{Noise Initialization Analysis.}
We compare retrieved initialization with five fixed Gaussian seeds to examine how noise selection affects editing performance.
Table~\ref{tab:fixed_noise} shows relatively small variations across the tested seeds. Although individual fixed seeds achieve better scores, retrieval remains competitive across the reported metrics. These results suggest that prompt-conditioned noise selection maintains overall editing performance and offers modest gains over the tested seed average, with a trade-off in layout preservation.

\section{Conclusion}
\label{sec:con}
This paper presented \method{}, a training free framework for multi subject video editing that jointly controls target localization, occlusion geometry, and appearance initialization. By combining mask-guided localization, depth-aware occlusion control, and noise latent prompting, \method{} enables precise target editing while preserving neighboring subjects and background content. Experiments on MSVBench show that \method{} achieves the lowest mean \warperr{} and \cmerr{} and the highest Q Edit among the evaluated methods, while maintaining competitive \clipt{} and \clipf{} scores. These results demonstrate the effectiveness of coordinated spatial, geometric, and latent guidance for controllable multi subject video editing.

\vfill\pagebreak

\bibliographystyle{IEEEbib}
\bibliography{strings,refs}

\end{document}